\documentclass{article}

\usepackage[square,numbers]{natbib}
\usepackage{authblk}
\usepackage[margin=1in]{geometry} 

\usepackage{amsmath}
\usepackage{amssymb}
\usepackage{amsthm}
\usepackage{mathtools}
\usepackage{optidef}
\usepackage{listings}
\usepackage{fullpage}
\usepackage{pifont}
\usepackage[utf8]{inputenc}
\usepackage[shortlabels]{enumitem}
\usepackage[colorlinks=true, urlcolor=blue]{hyperref}
\usepackage{cleveref}
\usepackage{xspace}
\usepackage{booktabs}
\usepackage{algorithm}
\usepackage{wasysym}
\usepackage[noend]{algpseudocode}
\usepackage{tikz}
\usepackage{enumitem}
\usepackage{pgfplots}
\usepackage{nicefrac}
\usepackage{bm}

\newtheorem{theorem}{Theorem}

\ifdefined\U
\renewcommand{\U}{\mathcal{U}}
\else
\newcommand{\U}{\mathcal{U}}
\fi

\newcommand{\A}{\mathcal{A}}
\newcommand{\R}{\mathbb{R}}
\newcommand{\D}{\mathcal{D}}
\newcommand{\N}{\mathcal{N}}

\DeclareMathOperator{\argmax}{argmax}

\begin{document}

\title{Graph-based Methods for Discrete Choice\thanks{This is the arXiv version of a paper published in \textit{Network Science}; see \url{https://doi.org/10.1017/nws.2023.20}.}}

\author{Kiran Tomlinson\thanks{kt@cs.cornell.edu} } 
\author{Austin R.\ Benson}
\affil{Department of Computer Science, Cornell University}
\date{}

\maketitle

\begin{abstract}
Choices made by individuals have widespread impacts---for instance, people choose between political candidates to vote for, between social media posts to share, and between brands to purchase---moreover, data on these choices are increasingly abundant. \emph{Discrete choice models} are a key tool for learning individual preferences from such data. Additionally, social factors like conformity and contagion influence individual choice. Traditional methods for incorporating these factors into choice models do not account for the entire social network and require hand-crafted features. To overcome these limitations, we use graph learning to study choice in networked contexts. We identify three ways in which graph learning techniques can be used for discrete choice: learning chooser representations, regularizing choice model parameters, and directly constructing predictions from a network. We design methods in each category and test them on real-world choice datasets, including county-level 2016 US election results and Android app installation and usage data. We show that incorporating social network structure can improve the predictions of the standard econometric choice model, the multinomial logit. We provide evidence that app installations are influenced by social context, but we find no such effect on app usage among the same participants, which instead is habit-driven. In the election data, we highlight the additional insights a discrete choice framework provides over classification or regression, the typical approaches.  On synthetic data, we demonstrate the sample complexity benefit of using social information in choice models. 
\end{abstract}

\section{Introduction}

Predicting and understanding the decisions individuals make has a host of applications, including modeling online shopping preferences~\citep{ruiz2020shopper}, forecasting the demand for renewable energy~\citep{axsen2012social, michelsen2012homeowners}, and analyzing elections~\citep{dreher2014determinants, glasgow2001mixed}. These and other scenarios are studied under the umbrella of \emph{discrete choice}~\citep{train2009discrete}, which describes any setting where people select items from a set of available alternatives. While discrete choice has its roots in econometrics, machine learning approaches have recently found great success in discrete choice applications~\citep{seshadri2019discovering,rosenfeld2020predicting,tomlinson2021learning,bower2020preference}.
This recent interest is driven by the increasing importance of Web-based choices (e.g., purchases on Amazon or bookings on Expedia), which provide both motivating applications and benchmark datasets. These new methods extend existing econometric models---most notably the classic \emph{conditional} or \emph{multinomial logit} (CL/MNL)~\citep{mcfadden1973conditional}---by learning more complex effects, such as context-dependent and non-linear preferences.

One of the crucial aspects of human decision-making is that, as fundamentally social creatures, our preferences are strongly influenced by our social context. Viral trends, conformity, word-of-mouth, and signaling all play roles in behavior, including choices~\citep{feinberg2020choices, axsen2012social}. Additionally, people with similar preferences, beliefs, and identities are more likely to be friends in the first place, a phenomenon known as \emph{homophily}~\citep{mcpherson2001birds}. Together, these factors indicate that social network structure could be very informative in predicting choices. In economics and sociology, there has been growing interest in incorporating social factors into discrete choice models~\citep{mcfadden2010sociality,maness2015generalized,feinberg2020choices}. However, the methods used so far in these fields have largely been limited to simple feature-based summaries of social influence (e.g., what fraction of someone's friends have selected an item~\citep{goetzke2011bicycle}). 

On the other hand, the machine learning community has developed a rich assortment of graph learning techniques that can incorporate entire social networks into predictive models~\citep{kipf2017semi,jia2022unifying,wu2020comprehensive}, such as graph neural networks and graph-based regularization. These approaches can handle longer-range interactions and are less reliant on hand-crafted features. Because of the large gulf between the discrete choice and machine learning communities, there has been almost no study of the application of graph learning methods to discrete choice, where they have the potential for major impact. Perhaps one factor hindering the use of graph learning in discrete choice is that machine learning methods are typically designed for either regression or classification. Discrete choice has several features distinguishing it from multiclass classification (its closest analogue)---for instance, each observation can have a different set of available items. As a concrete example, any image could be labeled as a cat in a classification setting, but people choosing between doctors may have their options dictated by their insurance policy.

Motivated by this need, we adapt graph learning techniques to incorporate social network structure into discrete choice modeling. By taking advantage of phenomena like homophily and social contagion, these approaches improve the performance of choice prediction in a social context. In particular, we demonstrate how graph neural networks can be applied to discrete choice, derive Laplacian regularization for the multinomial logit model, and adapt label propagation for discrete choice. We show in synthetic data that Laplacian regularization can improve sample complexity by orders of magnitude in an idealized scenario. 

To evaluate our methods, we perform experiments on real-world election data and Android app installations, with networks derived from Facebook friendships, geographic adjacency, and Bluetooth pings between phones. We find that such network structures can improve the predictions of discrete choice models in a semi-supervised learning task. For instance, Laplacian regularization improves the mean relative rank (MRR)\footnote{MRR measures the relative position of the true choice in the predicted ranking~\citep{tomlinson2021learning}.} of predictions by up to a 6.8\% in the Android app installation data and up to 2.6\% in the 2016 US election data. In contrast with our results on app installations, we find no evidence of social influence in app usage among the same participants: social factors appear to influence the apps people get, but less so the apps they actually use. Instead, we find that app usage is dominated by personal habit. Another interesting insight provided by our discrete choice models in the app installation data is the discovery of two separate groups of participants, one in which Facebook is popular, while the other prefers Myspace.\footnote{The dataset is from 2010, when both were popular options.} We further showcase the power provided by a discrete choice approach by making counterfactual predictions in the 2016 US election data with different third-party candidates on the ballot. While a common narrative is that Clinton's loss was due to spoiler effects by third-party candidates~\citep{guardian,rollcall}, our results do not support this theory, although we emphasize the likelihood of confounding factors. Our tools enable us to rigorously analyze these types of questions.

\section{Related work}

There is a long line of work in sociology and network science on social behavior, including effects like contagion and herding~\citep{centola2007complex,easley2010networks,banerjee1992simple}. More recently, there has been interest in the use of discrete choice in conjunction with network-based analysis~\citep{feinberg2020choices} enabled by rich data with both social and choice components~\citep{aharony2011social}. The traditional econometric approach to discrete choice modeling with social effects is to add terms to an individual's utility that depend on the actions or preferences of others~\citep{brock2001discrete,mcfadden2010sociality,maness2015generalized}. For instance, this approach can account for an individual's desire for conformity~\citep{bernheim1994theory}. This is done by treating the choices made by a chooser's community as a feature of the chooser and applying a standard multinomial logit~\citep{paez2008discrete, kim2014expanding, goetzke2011bicycle, walker2011correcting, kim2018social}. In contrast, we focus on methods that employ the entire graph rather than derived features. This enables methods to account for longer-range interactions and phenomena such as network clustering without hand-crafting features. We are aware of one econometric paper that uses preference correlations over a full network in a choice model~\citep{leung2019inference}, but inference under this method requires Monte Carlo simulation. Laplacian regularization, on the other hand, allows us to find our model's maximum likelihood estimator with straightforward convex optimization. Mixture models are another way of incorporating structured preference heterogeneity into discrete choice, such as the mixed logit~\citep{mcfadden2000mixed} and hierarchical Bayes models with mixture priors~\citep{allenby2006hierarchical,burda2008bayesian}. Again, these approaches present significant challenges for inference, requiring Monte Carlo methods, variational approximations, or expectation maximization. Additionally, in positing unknown latent populations, mixture models ignore the key information provided by the structure of the network. Another large area of research in discrete choice concerns models that allow deviations from the axiom of \emph{independence of irrelevant alternatives} (IIA)~\citep{luce1959individual}. Many of these models, such as the multinomial probit~\citep{hausman1978conditional}, are very challenging to estimate. To keep our focus on incorporating network effects, we use tractable logit models obeying IIA. However, there are recent non-IIA models admitting efficient inference to which we could apply our methods~\citep{seshadri2019discovering,bower2020preference,tomlinson2021learning}; this is beyond the scope of the present work, but we expand further on this idea in the discussion.

In another direction, there are many machine learning methods that use network structure in predictive tasks; graph neural networks (GNNs)~\citep{kipf2017semi, xu2018powerful, wu2020comprehensive} are a popular example. Discrete choice is related to classification tasks, but the set of available items (i.e., labels) is specific to each observation---additionally, discrete choice models are heavily informed by economic notions of preference and rationality~\citep{mcfadden1973conditional,train2009discrete}. A more traditional machine learning method of exploiting network structure for classification is label propagation~\citep{xiaojin2002learning}, which we extend to the discrete choice setting. Recent work has shown how to combine label propagation with GNNs for improved performance~\citep{jia2020residual} and presented a unified generative model framework for label propagation, GNNs, and Laplacian regularization~\citep{jia2022unifying}. The present work can be seen as an adaptation and empirical study of the methods from~\citep{jia2022unifying} for discrete choice rather than regression. 

The idea of applying Laplacian regularization to discrete choice models appeared several years ago in an unpublished draft~\citep{zhang2017social}. However, the draft did not provide experiments beyond binary choices (which reduces to standard semi-supervised node classification~\citep{kipf2017semi}). In contrast, we compare Laplacian regularization with other methods of incorporating social network structure (GNNs and propagation) on real-world multi-alternative choice datasets.

There is a large body of existing research on predicting app usage and installation, including using social network structure~\citep{baeza2015predicting,pan2011composite,xu2013preference}, but our use of network-based discrete choice models for this problem is novel. Our approach has the advantage of being applicable to both usage and installation with minimal differences, allowing us to compare the relative importance of social structure in these settings. Other lines of related work apply discrete choice models to networks in order to model edge formation~\citep{overgoor2019choosing,tomlinson2021learning,gupta2022mixed,overgoor2020scaling}, or jointly model network dynamics and the behavior of nodes~\cite{snijders2010introduction,steglich2010dynamic}. In contrast, we fix a network and use it as a signal to improve choice prediction, without addressing network dynamics.

\section{Preliminaries}\label{sec:preliminaries}
In a discrete choice setting, we have a universe of \emph{items} $\U$ and a set of \emph{choosers} $\A$. In each choice instance, a chooser $a\in \A$ observes a \emph{choice set} $C \subset \U$ and chooses one item $i \in C$. Each item $i\in \U$ may be described by a vector of features $\bm{y}_i \in \R^{d_y}$. Similarly, a chooser $a$ may have a vector of features $\bm{x}_a \in \R^{d_x}$. In the most general form, a choice model assigns choice probabilities for $a$ to each item $i\in C$: 

\begin{equation}\label{eq:choice-model}
  \Pr(i \mid a, C) = \frac{\exp(u_{\bm{\theta}}(i, C, a))}{\sum_{j \in C} \exp(u_{\bm{\theta}}(i, C, a))},
\end{equation}
where $u_{\bm{\theta}}(i, C, a)$ is the utility of item $i$ to chooser $a$ when seen in choice set $C$, a function with parameters ${\bm{\theta}}$. Note that since the utilities in \Cref{eq:choice-model} can depend on the choice set, this general form can express choice probabilities that vary arbitrarily across choice sets (this is sometimes called the \emph{universal logit}). When constructing more useful parsimonious models, the utilities $u_{\bm{\theta}}(i, C, a)$ can depend on $\bm{x}_a$, $\bm{y}_i$, both, or neither. In the simplest case---the traditional logit model---$u_{\bm{\theta}}(i, C, a) = u_i$ is constant over choosers and sets. This formulation is attractive from an econometric perspective, since it corresponds to a rationality assumption: if we suppose a chooser has underlying utilities $u_1, \dots, u_k$ and observes a perturbation of their utilities $u_i + \varepsilon_i$ (where $\varepsilon_i$ follows a Gumbel distribution)  before selecting the maximum observed utility item, then their resulting choice probabilities take the form of a logit~\citep{mcfadden1973conditional}. 

When we add a linear term in chooser features to the logit model, the result is the \emph{multinomial logit} (MNL)~\citep{hoffman1988multinomial, mcfadden1973conditional}, with utilities $u_{\theta}(i, C, a) = u_i + \bm{\gamma_i}^T\bm{x_a}$, where $u_i$ are item-specific utilities and $\bm{\gamma_i}$ is a vector of item-specific coefficients capturing interactions with the chooser features $\bm{x_a}$. Similarly, when we add a linear term in item features, the result is a \emph{conditional logit} (CL), with utilities $u_i +  \bm{\varphi}^T\bm{y_i}$. The \emph{conditional multinomial logit} (CML) has both the chooser and item feature terms: $u_i +  \bm{\varphi}^T\bm{y_i} + \bm{\gamma_i}^T\bm{x_a}$. In order to capture heterogeneous preferences among a group of choosers, one natural approach is to allow each chooser $a$ to have different logit utilities. We call this a \emph{per-chooser logit}, which is specified by per-chooser utilities $u_{\bm \theta}(i, C, a) = u_{ia}$. Similarly, a \emph{per-chooser conditional logit} has varying item feature coefficients $\bm \varphi_a$ for each chooser $a$, with $u_{\bm \theta}(i, C, a) = u_{ia} + \bm{\varphi}_a^T\bm{y_i}$. More generally, we call any choice model parameter which varies across choosers a \emph{per-chooser parameter}. 

In addition to this standard discrete choice setup, our settings also have a network describing the relationships between choosers. Choosers are nodes in an undirected graph $G = (\A, E)$ where the presence of an edge $(a, b)\in E$ indicates a connection between $a$ and $b$ (e.g., a friendship). We assume $G$ is connected. The \emph{Laplacian} of $G$ is $L = D - A$, where $D$ is the diagonal degree matrix of $G$ and $A$ is the adjacency matrix. The Laplacian has a number of useful applications, including in graph clustering~\citep{hagen1991fast} and counting spanning trees~\citep{merris1994laplacian}. For our purposes, the key property of the Laplacian is that quadratic forms of $L$ measure how much a node-wise vector differs across edges of the graph (we elaborate on this property below). We use $n = |\A|$, $m = |E|$, and $k = |U|$. Finally, $I$ denotes the identity matrix.

\section{Graph-based methods for discrete choice}\label{sec:methods}
We identify three phases in choice prediction where networks can be incorporated: networks can be used (1) to inform model parameters, (2) to learn chooser representations, or (3) to directly produce predictions. In this section, we develop representative methods in each category. We briefly describe each method before diving into more detail.

First, networks can inform inference for a model that already accounts for chooser heterogeneity. This is done by incorporating the correlations in utilities (or other choice model parameters) of individuals who are close to each other in the network; we refer to these as \emph{preference correlations} for simplicity. Our Laplacian regularization approach (described in~\Cref{sec:laplacian}) does exactly this, and we show that it corresponds to a Bayesian prior on network-based preference correlations. Second, networks can be used to learn latent representations of choosers that are then used as features in a choice model like the MNL. GNNs have been extensively studied as representation-learning tools---in \Cref{sec:gcn}, we focus on how to incorporate them into choice models, using graph convolutional networks (GCNs)~\citep{kipf2017semi} as our canonical example. Third, direct network-based methods (such as label propagation~\citep{xiaojin2002learning}, which repeatedly averages a node's neighboring labels) can also be used as a simple baseline for choice predictions. While this approach is simple and efficient, it lacks the proper handling of choice sets of the previous probabilistic approaches. Nonetheless, we find it a useful and effective baseline, and we adapt label propagation for discrete choice in \Cref{sec:propagation}.

\subsection{Laplacian regularization}\label{sec:laplacian}
We begin by describing how to incorporate network information in a choice model like MNL through Laplacian regularization~\citep{ando2007learning}. Laplacian regularization encourages parameters corresponding to connected nodes to be similar through a loss term of the form $\lambda\bm{\alpha}^TL \bm{\alpha}$, where $L$ is the graph Laplacian (as defined in \Cref{sec:preliminaries}), $\bm{\alpha}$ is the vector of parameter values for each node, and $\lambda$ is the scalar regularization strength. A famous identity is that $\bm{\alpha}^TL \bm{\alpha} = \sum_{(i, j) \in E} (\bm{\alpha}_i - \bm{\alpha}_j)^2$, which more clearly shows the regularization of connected nodes' parameters towards each other. This also shows that the Laplacian is positive semi-definite, since $\bm{\alpha}^TL \bm{\alpha} \ge 0$, which will be useful to preserve the convexity of the multinomial logit's (negative) log-likelihood.

The idea of using Laplacian regularization for discrete choice was proposed in~\citep{zhang2017social} (although they focused on regularizing intercept terms in binary logistic regression). We generalize the idea to be applicable to any logit-based choice model and show that it corresponds to Bayesian inference with a network correlation prior. We then specialize to the models we use in our experiments. Laplacian regularization is simple to implement, can be added to any logit-based choice model with per-chooser parameters, and only requires training one extra hyperparameter. Laplacian regularization also carries a number of advantages over another approach to accounting for structured preference heterogeneity, mixture modeling.

\subsubsection{Theory of Laplacian-regularized choice models}
Consider a general choice model, as in \Cref{eq:choice-model}. We split the parameters $\bm{\theta}$ into two sets $\theta_\A$ and $\theta_{G}$, where parameters $\bm{\alpha} \in \theta_\A, \bm{\alpha}\in \R^n$ vary over choosers and parameters $\beta \in \theta_{G}, \beta \in \R$ are  constant over choosers. The log-likelihood of a general choice model is:
\begin{equation}
  \ell(\bm{\theta}; \D) = \sum_{(i, a, C) \in \D} \Big[\log(u_{\bm{\theta}}(i, C, a)) - \log \sum_{j \in C} \exp(u_{\bm{\theta}}(j, C, a))\Big].\label{eq:chooice-model-ll}
\end{equation}
The Laplacian- and $L_2$-regularized log-likelihood (with $L_2$ regularization strength $\gamma$) is then
\begin{align}
  \ell_L(\bm{\theta}; \D) &= \ell(\bm{\theta}; \D) - \frac{\lambda}{2}\sum_{\bm{\alpha} \in \theta_\A} \bm{\alpha}^T L \bm{\alpha} -\frac{\gamma}{2} \sum_{\bm{\alpha} \in \theta_\A} ||\bm{\alpha}||_2^2.\label{eq:laplacian-reg-ll}
\end{align}
We show that regularized maximum likelihood estimation of $\bm{\theta}$ corresponds to Bayesian inference with a prior on per-chooser parameters that encourages smoothness over the network. In contrast, existing results on priors for semi-supervised regression~\citep{xu2010empirical, chin2019decoupled} typically split the nodes into observed and unobserved, fixing the observed values and only considering randomness over unobserved nodes. In choice modeling, observing  choices at a node only updates our beliefs about their preferences, leaving some uncertainty. Our result also allows some parameters of the choice model to be chooser-dependent and others to be constant across choosers, allowing it to be fully general over choice models. Finally, we  note that $L_2$ regularization can also be applied to the global parameters $\beta$, which as usual corresponds to a Gaussian prior on these parameters---however, we state the result with uniform priors to emphasize the Laplacian regularization on the per-chooser parameters $\bm \alpha$. 


\begin{theorem}\label{thm:bayesian}
  The maximizer $\bm{\theta}^*_{\text{MLE}}$ of the Laplacian- and $L_2$-regularized log-likelihood $\ell_L(\bm{\theta}; \D)$ is the maximum a posteriori estimate $\bm{\theta}^*_{\text{MAP}}$ after observing $\D$ under the i.i.d.\ priors $\bm{\alpha} \sim \N(0,  [\lambda L + \gamma I] ^{-1})$ for each $\bm{\alpha} \in \theta_\A$ and i.i.d.\ uniform priors for each $\beta \in \theta_{G}$.
\end{theorem}

\begin{proof}
First, recall that $L$ is positive semi-definite, so $\lambda L + \gamma I $ (with $\gamma, \lambda > 0$) is positive definite and invertible. Now, using Bayes' Theorem,
\begin{align*}
\bm{\theta}^*_{\text{MAP}} &= \argmax_{\bm{\theta}} p(\bm{\theta} \mid \D)\\
&= \argmax_{\bm{\theta}} \frac{\Pr(\D \mid \bm{\theta}) p(\bm{\theta})}{\Pr(\D)}.
\end{align*}   
Since $\Pr(\D)$ is independent of the parameters and $\log$ is monotonic and increasing,
\begin{align*}
 \bm{\theta}^*_{\text{MAP}} &= \argmax_{\bm{\theta}} \left[\log \Pr(\D \mid \bm{\theta})+ \log p(\bm{\theta})\right].
\end{align*}
Notice that the first term is exactly the log-likelihood $\ell(\bm{\theta}; \D)$. Additionally, the priors of each parameter are independent, so
\begin{align*}
  \log p(\bm{\theta}) &= \sum_{\bm{\alpha} \in \theta_\A} \log p(\bm{\alpha}) +  \sum_{\beta \in \theta_{G}} \log p(\beta).
\end{align*}
Since the priors $p(\beta)$ are uniform, they do not affect the maximizer:
\begin{align*}
  \bm{\theta}^*_{\text{MAP}} &= \argmax_{\bm{\theta}} \left[\ell(\bm{\theta}; \D)+ \sum_{\bm{\alpha} \in \theta_\A} \log p(\bm{\alpha})\right].
\end{align*}
Now consider the Gaussian priors $p(\bm{\alpha})$:
\begin{align*}
  p(\bm{\alpha}) &= (2\pi)^{n/2} \det[(\lambda L + \gamma I) ^{-1}]^{-1/2} \exp\left(-\frac{1}{2} \bm{\alpha}^T [(\lambda L + \gamma I) ^{-1}]^{-1}\bm{\alpha}\right).
\end{align*}
Simplifying the term in the $\exp$ reveals the two regularization terms:
\begin{align*}
-\frac{1}{2} \bm{\alpha}^T [(\lambda L + \gamma I) ^{-1}]^{-1}\bm{\alpha}&=-\frac{1}{2} \bm{\alpha}^T (\lambda L+\gamma I) \bm{\alpha}\\
  &= -\frac{\lambda}{2} \bm{\alpha}^T L\bm{\alpha} -\frac{\gamma}{2} \bm{\alpha}^T\bm{\alpha} \\
   &= -\frac{\lambda}{2} \bm{\alpha}^T L\bm{\alpha} -\frac{\gamma}{2} ||\bm{\alpha}||_2^2. 
\end{align*}
We thus have, for a constant $c$ independent of $\bm \alpha$,
\begin{align*}
  \log p(\bm \alpha) &= \log \left((2\pi)^{n/2} \det[(\lambda L + \gamma I) ^{-1}]^{-1/2} \exp\left(-\frac{\lambda}{2} \bm{\alpha}^T L\bm{\alpha} -\frac{\gamma}{2} ||\bm{\alpha}||_2^2 \right)\right)\\
  &= -\frac{\lambda}{2} \bm{\alpha}^T L\bm{\alpha} -\frac{\gamma}{2} ||\bm{\alpha}||_2^2 + c.
\end{align*}
Plugging this is in and dropping the constants not affecting the maximizer yields
\begin{align*}
  \bm{\theta}^*_{\text{MAP}} &= \argmax_{\bm{\theta}} \left[\ell(\bm{\theta}; \D) - \frac{\lambda}{2} \sum_{\bm{\alpha} \in \theta_\A}  \bm{\alpha}^T L \bm{\alpha}  -\frac{\gamma}{2} \sum_{\bm{\alpha} \in \theta_\A} ||\bm{\alpha}||_2^2 \right]\\
  &= \argmax_{\bm{\theta}} \ell_L(\bm{\theta}; \D)\\
  &= \bm{\theta}^*_{\text{MLE}}.\qedhere
\end{align*}
\end{proof}
Notice that the Gaussian in the theorem above has precision (i.e., inverse covariance) matrix $\lambda L + \gamma I$.  The partial correlation between the per-chooser parameters $\bm{\alpha}_i$ and $\bm{\alpha}_j$, $i \ne j$, (controlling for all other nodes) is therefore
\begin{equation}
  - \frac{\lambda L_{ij}}{\sqrt{(\lambda L_{ii}  + \gamma) (\lambda L_{jj} + \gamma)}} = \frac{\lambda A_{ij}}{\sqrt{(\lambda d_i  + \gamma) (\lambda d_j + \gamma)}}
\end{equation}
using the standard Gaussian identity relating precision and partial correlation~\citep{liang2015equivalent} (where $d_i$ is the degree of $i$). If both $d_i, d_j > 0$ and $\gamma$ is small, then we can approximate
\begin{equation}
  \frac{\lambda A_{ij}}{\sqrt{(\lambda d_i  + \gamma) (\lambda d_j + \gamma)}} \approx \frac{\lambda A_{ij}}{\sqrt{(\lambda d_i) (\lambda d_j)}}  = \frac{A_{ij}}{\sqrt{d_i d_j}}.
\end{equation}
This is easy to interpret: $\bm{\alpha}_i$ and $\bm{\alpha}_j$ have partial correlation 0 when $i$ and $j$ are unconnected ($A_{ij} = 0$) and positive partial correlation when they are connected (larger when they have fewer other neighbors). That is, the Gaussian prior in the theorem assumes neighboring nodes have correlated preferences.

\subsubsection{Laplacian-regularized logit models} To incorporate Laplacian regularization in our four logit models (logit, MNL, CL, CML), we add per-chooser utilities $v_{ia}$ for each item $i$ and chooser $a$ to the utility formulations. For instance, this results in the following utility function for a per-chooser MNL: $u_\theta(i, C, a) = u_i + \bm{x_a}^T\bm{\gamma_i} + v_{ia}$. While we could get rid of the global utilities $u_i$, $L_2$ regularization enables us to learn a parsimonious model where $u_i$ is the global baseline utility and $v_{ia}$ represents per-chooser deviations. The per-chooser parameters of a Laplacian-regularized logit are $\theta_A = \{\bm{v_i}\}_{i\in U}$, where the vector $\bm{v_i}$ stacks the values of $v_{ia}$ for each chooser $a\in A$. All other parameters are global. The Laplacian- and $L_2$-regularized log-likelihood can then be written down by combining \Cref{eq:chooice-model-ll,eq:laplacian-reg-ll}. Crucially, since the Laplacian is positive semi-definite, the terms $-\frac{\lambda}{2} \bm{v_i}^T L \bm{v_i}$ are concave---and since all four logit log-likelihoods are concave (as is the $L_2$ regularization term), their regularized negative log likelihoods (NLLs) are convex. This enables us to easily learn maximum-likelihood models with standard convex optimization methods.

\subsection{Graph neural networks}\label{sec:gcn}
Graph neural networks (GNNs)~\citep{wu2020comprehensive} use a graph to structure the aggregations performed by a neural network, allowing parameters for neighboring nodes to influence each other. We test the canonical GNN, a graph convolutional network (GCN)~\citep{kipf2017semi}, where node embeddings are averaged across neighbors before each neural network layer. There are many other types of GNNs (see~\citep{wu2020comprehensive} for a survey)---we emphasize that we do not claim this particular GCN approach to be optimal for discrete choice. Rather, we illustrate how GNNs can be applied to choice data and encourage further exploration. 

In a depth-$d$ GCN, each layer performs the following operation, producing a sequence of embeddings $H^{(0)}, \dots, H^{(d)}$:
\begin{equation}
  H^{(i+1)} = \sigma(A'H^{(i)}W^{(i)})
\end{equation} 
where $H^{(0)}$ is initialized using node features (if they are available---if not, $H^{(0)}$ is learned), $\sigma$ is an activation function, $W^{(i)}$ are parameters, and $A' = (D+2I)^{-\nicefrac{1}{2}}(A+I)(D+2I)^{-\nicefrac{1}{2}}$ is the degree-normalized adjacency matrix (with self-loops). Self-loops are added to $G$ to allow a node's current embedding to influence its embedding in the next layer. We can either use $H^{(d)}$ as the final embeddings or concatenate each layer's embedding into a final embedding $H$. In our experiments, we use a two-layer GCN (both with output dimension 16) and concatenate the layer embeddings. For simplicity, we fix the dropout rate at 0.5. 

To apply GCNs to discrete choice, we can treat the final node embeddings as chooser features and apply an MNL, modeling utilities as $u_\theta(i, C, a) = u_i + H_a^T\bm{\gamma_i}$, where $u_i$ and $\bm{\gamma_i}$ are per-item parameters (the intercept and embedding coefficients, respectively). If item features are also available, we add the conditional logit term $\bm{\theta}^T \bm{y_i}$.  Thanks to automatic differentiation software such as PyTorch~\citep{paszke2019pytorch}, we can train both the GCN and MNL/CML weights end-to-end. Again, any node representation learning method could be used for the embeddings $H$---we use a GCN for simplicity. 

In general, graph neural networks have the advantage of being highly flexible, able to capture complex interactions between the features of neighboring nodes. However some recent research has indicated that non-linearity is less helpful for classification in GNNs than in traditional neural networks tasks~\citep{wu2019simplifying}. With the additional modeling power comes significant additional difficulty in training and hyperparameter selection (for embedding dimensions, depth, dropout rate, and activation function).

\subsection{Choice fraction propagation}\label{sec:propagation}
We also consider a baseline method that uses the graph to directly derive choice predictions, without a probabilistic model of choice. We extend label propagation~\citep{zhou2004learning,jia2022unifying} to multi-alternative discrete choice. The three features distinguishing the choice setting from standard label propagation is that we can observe multiple ``labels'' (i.e., choices) per chooser, each observation may have had different available labels, and that not all labels are available at inference time. Given training data of observed choices of the form $(i, C, a)$, where chooser $a \in \A$ chose item $i\in C \subseteq \U$, we assign each chooser $a$ a vector $\bm{z_a}$ of size $k = |U|$ with each item's \emph{choice fraction}. That is, the $i$th entry of $\bm{z_a}$ stores the fraction of times $a$ chose $i$ in the observed data out of all opportunities they had to do so (i.e., the number of times $i$ appeared in their choice set). We use choice fraction rather than counts to normalize by the number of observations for a chooser and not to count against an item instances when it was not available. 

We then apply label propagation to the vectors $\bm{z_a}$ over $G$. Let $Z^{(0)}$ be the matrix whose rows are $\bm{z_a}$. As in standard label propagation, we iterate
$
 Z^{(i+1)} \gets (1-\rho)  Z^{(0)} + \rho D^{-\nicefrac{1}{2}}AD^{-\nicefrac{1}{2}}  Z^{(i)}
$
until convergence, where $\rho \in [0, 1]$ is a hyperparameter that controls the strength of the smoothing. Let $ Z^{(\infty)}$ denote the stationary point of the iterated map. For inference, we can use the $a$th row of $Z^{(\infty)}$ (in practice, we will have a matrix arbitrarily close to $Z^{(\infty)}$), denoted $\bm{z_a^{(\infty)}}$, to make predictions for chooser $a$. Given a choice set $C$, we predict $a$ will choose the argmax of $\bm{z_a^{(\infty)}}$ restricted to items appearing in $C$. Note that in a semi-supervised setting, we do not observe any choices from the test choosers, so their entries of $Z^{(0)}$ will be zero. The term $(1-\rho)Z^{(0)}$ then acts as a uniform prior, regularizing the test chooser entries of $Z^{(\infty)}$ towards 0. Since choice fraction propagation does not use chooser or item features, it is best suited to scenarios where neither are available.

\section{Networked discrete choice data}\label{sec:data}
We now describe several datasets in which we can leverage social network structure for improved choice prediction using the methods we develop. \Cref{tab:datasets} shows a summary of our datasets, which are available at \url{https://osf.io/egj4q/}.

\begin{table}
\centering
 \caption{Dataset summary. $|A|$: number of choosers (aggregated at the county/precinct for elections), $|U|$: number of items, $|C|$: choice set sizes, $N$: number of observed choices, $d_x$: number of chooser features.}  \label{tab:datasets}
  \begin{tabular}{lrrrrrr}
  \toprule
  Dataset & \bfseries{$|A|$} & \bfseries{$|U|$} & \bfseries{$|C|$} & \bfseries{$N$} & $d_x$ & $d_y$\\
  \midrule
  \textsc{app-install} & 139 & 127 & 51--127 & 4,039 & --- & --- \\ 
  \textsc{app-usage} & 104 & 121 & 2--55 & 20,564 & --- & 1\\ 
  \textsc{us-election-2016} & 3,112 & 32 & 3--22 & 135,382,576 & 19 & ---\\ 
  \textsc{ca-election-2016} & 21,495 & 182 & 2 & 261,278,336$^*$ &  17 & ---\\ 
  \textsc{ca-election-2020} & 17,282 & 170 & 2 & 225,606,176$^*$ & 17 & ---\\ 
  \bottomrule
    \multicolumn{4}{l}{\footnotesize{$^*$Voters had more than one election (i.e., choice) on their ballots.}}
\end{tabular}
\end{table}

\subsection{Friends and Family app data}\label{sec:app-data}

The Friends and Family dataset~\citep{aharony2011social} follows over 400 residents of a young-family community in North America during 2010-2011. The dataset is remarkably rich, capturing many aspects of the participants' lives. For instance, they were given Android phones with purpose-made logging software that captured app installation and usage as well as Bluetooth pings between participants' phones. We use the installation and usage data to construct two separate choice datasets (\textsc{app-install} and \textsc{app-usage}) and use a network built from Bluetooth pings, as in~\citep{aharony2011social}. We ignore uncommon and built-in apps (for instance, we ignore apps whose package names begin with \texttt{com.android}, \texttt{com.motorola}, \texttt{com.htc}, \texttt{com.sec}, and \texttt{com.google}), leaving a universe $\U$ of 127 apps in \textsc{app-install} and 121 in \textsc{app-usage} (e.g., Twitter, Facebook, and Myspace).

To construct \textsc{app-install}, we use scans that checked which apps were installed on each participant's phone every 10 minutes -- 3 hours. Each time a new app $i$ appears in a scan for a participant $a$, we consider that a choice from the set of apps $C$ that were not installed at the time of the last scan. We use a plain logit as the baseline model in \textsc{app-install}, since no item features are readily available. To construct \textsc{app-usage}, we use 30-second resolution scans of running apps. To separate usage into sessions, we select instances where a participant ran an app for the first time in the last hour. We consider such app runs to be a choice $i$ from the set of all apps $C$ installed on participant $a$'s phone at that time. Our discrete choice approach enables us to account for these differences in app availability. In \textsc{app-usage}, we use a conditional logit with a single instance-specific item feature: \emph{recency}, defined as $\log^{-1}(\text{seconds since last use})$ or 0 if the user has not used the app. While it would be possible to construct more complex sets of features with additional effort (for instance categorizing different types of apps or tracking down their Android store ratings), a simple baseline suffices to demonstrate how social network structure can benefit choice modeling even in the absence of item and user features.

\begin{figure}
\centering
  \includegraphics[width=0.8\columnwidth]{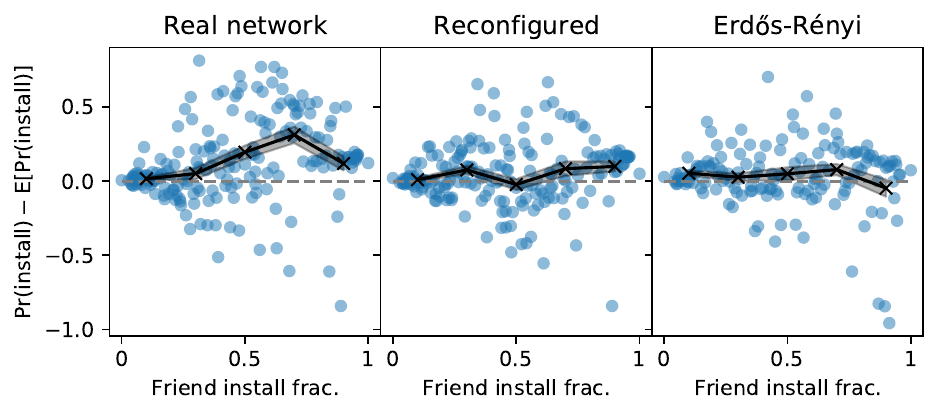}
  \caption{Difference between actual and expected install rates (if friendships were irrelevant). The left subplot is with the real network, while the right two are two null models. Each dot is a (participant, app) pair. The black line marks the mean over five bins, with the shaded region showing the standard error of the mean.}\label{fig:install_rates}
\end{figure}

 To form the social network $G$ over participants for both datasets, we use Bluetooth proximity hits---like the original study~\citep{aharony2011social}, we only consider hits in the month of April between 7am and midnight (to avoid coincidental hits from neighbors at night). For each participant $a$, we form the link $(a, b)$ to each of their 10 most common interaction partners $b$ (we also tested thresholds 2--9, but our methods all performed very similarly). We perform this thresholding because the Bluetooth ping network is extremely dense and contains many edges that are likely not socially meaningful (for instance, nearby phones may ping each other when two strangers shop in the same store). Prior research on this data found that social contacts were useful in predicting app installations, but did not employ a discrete choice approach~\citep{aharony2011social,pan2011composite}.  Our discrete choice approach allows us to account for multi-hop social connections and the context of each installation (i.e., what apps were already installed). 
 
As a warm-up data analysis, we show that people are more likely to install an app the more of their friends have it (but not if we randomize friendships). Let $n$ the total number of people, $n_i$ be the number of people who installed application $i$, $f_a$ the number of friends of person $a$, and $f_{ai}$ the number of friends of person $a$ who have app $i$. Suppose app installations are independent of friendships. If we sample some person $a$ uniformly at random and check which of their friends have app $i$, then the probability that $a$ also has app $i$ is $(n_i - f_{ai})/(n-f_a)$ (simply the remaining fraction of people who have the app, after observing the friends of $a$). However, if app installations correlate across friendships, the observed probability would be higher when $f_{ai} / f_a$ is larger. We measure the empirical probability that a person has an app at different friend-installation fractions. Specifically, we measure
\begin{equation}
  \frac{1}{nk} \sum_{i \in U, a\in \A} \left(\boldsymbol{1}_{ai} - \frac{n_i - f_{ai}}{n-f_a}\right),
\end{equation}
where $\boldsymbol{1}_{ai}$ is an indicator for whether person $a$ has app $i$. Notice that if friendships are uncorrelated with app installations, the expectation of the summand is $0$. Instead of taking the mean over all app pairs, we take the mean at each unique friend-installation fraction to see if having more friends with an app results in stronger deviations from uniform installations. This is exactly what we observe: when people have more friends with an app, they are more likely to install it (\Cref{fig:install_rates}). In contrast with two null models (a configuration model with the same degree distribution and an Erd\H{o}s--R\'{e}nyi graph with the same density), we see an increase in peoples' installation probabilities as a larger fraction of their friends have an app. This is in line with findings that the probability an individuals joins a social network community increases with the number of their friends in the community~\citep{backstrom2006group}. However, it is worth emphasizing that this finding is purely correlational---we have no way of knowing whether increased installation rates are due to homophily in the social network, word-of-mouth contagion, or other confounding factors.

\subsection{County-level US presidential election data}

\begin{figure}
  \centering
  \includegraphics[width=0.7\textwidth]{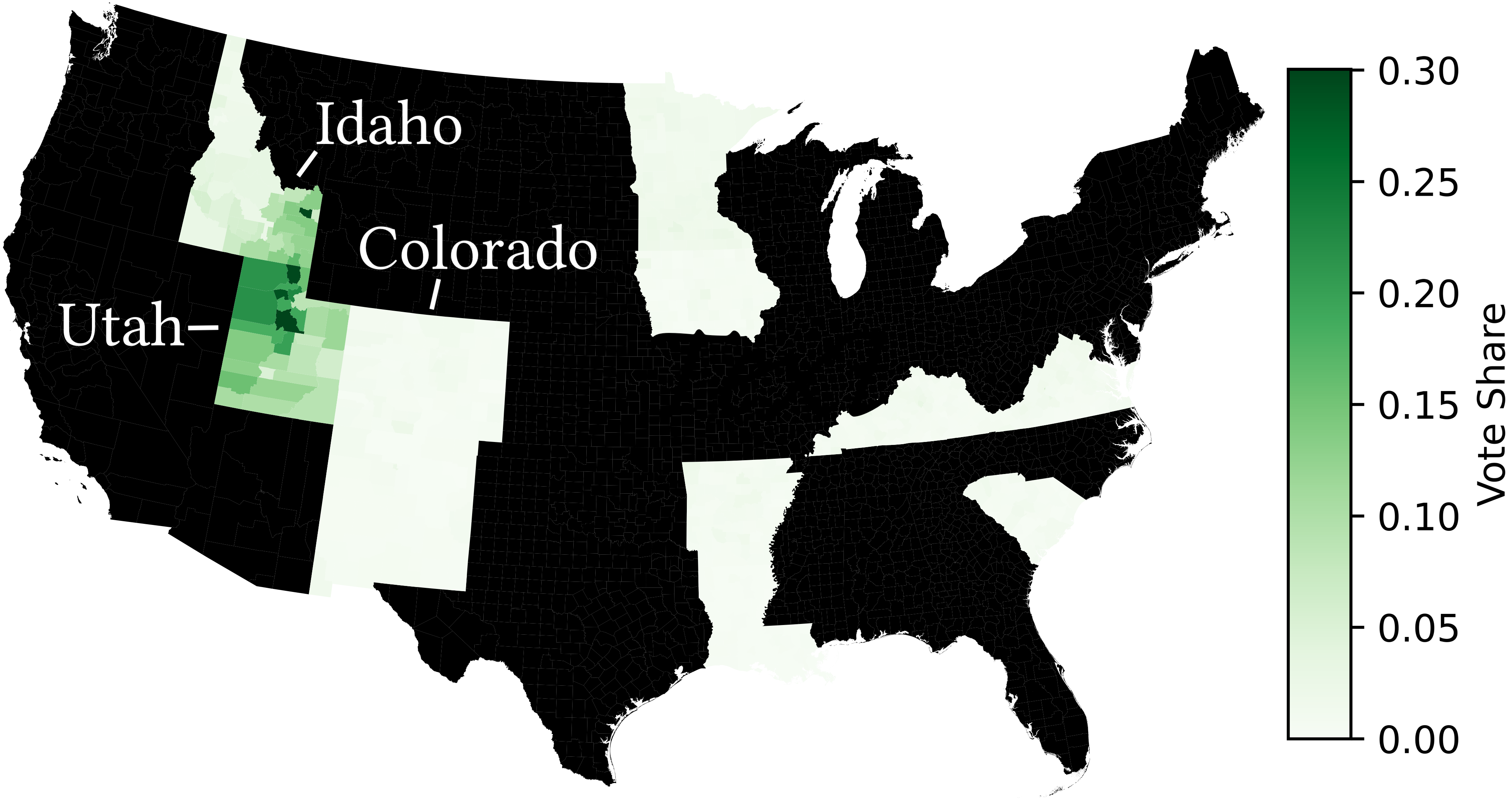}\\
  \caption{2016 US presidential election vote shares for conservative independent Evan McMullin. Notice his regional popularity and the spillover from Utah to southeast Idaho. McMullin was not on the ballot in filled-in states. The lack of spillover into Colorado may be due to its crowded field (22 candidates) or because it is less conservative than Idaho.}
  \label{fig:evan}
\end{figure}

US presidential election data is a common testbed for graph learning methods using a county-level adjacency network, but the typical approaches are to treat elections as binary classification or regression problems (predicting the vote shares of one party)~\citep{jia2020residual,zhou2021understanding,huang2020combining}. However, this ignores the fact that voters have more than two options---moreover, different candidates can be on the ballot in different states. The universe of items $\U$ in our 2016 election data contains no fewer than 31 different candidates (and a ``none of these candidates'' option in Nevada, which received nearly 4\% of the votes in one county). While third-party candidates are unlikely to win in the US, they often receive a non-trivial (and quite possibly consequential) fraction of votes. For instance, in the 2016 election, independent candidate Evan McMullin received 21.5\% of the vote in Utah, while Libertarian candidate Gary Johnson and Green Party candidate Jill Stein received 3.3\% and 1.1\% nationally (the gap between Clinton and Trump was only 2\%). A discrete choice approach enables us to include third-party candidates and account for different ballots in different states. As a visual example, in \Cref{fig:evan} we show the states in which McMullin appeared on the ballot as well as his per-county vote share. By accounting for ballot variation, we can make counterfactual predictions about what would happen if different candidates had been on the ballot, which is difficult without a discrete choice framework. For example, given McMullin's regional support in Utah, it is possible that he would have fared better in Nevada (Utah's western neighbor) than in an East Coast state like New York. Using the entire ballots also allows us to account for one possible reason why McMullin's vote share appears not to spilled over into Colorado, while it did into Idaho: Colorado had fully 22 candidates on the ballot, while Idaho only had 8. A discrete choice approach handles this issue cleanly, while regression on vote shares does not. We note that, due to inherent limitations of observational data, we cannot be sure of the causes of the effects we observe~\citep{tomlinson2021choice}---nonetheless, a discrete choice approach enables more flexible modeling and can improve prediction performance regardless of the cause of preference correlations. 

We gathered county-level 2016 presidential voting data from~\citep{kearney2016election} and county data from~\citep{jia2020residual},\footnote{One county---Oglala Lakota County, South Dakota (FIPS 46102)---was named Shannon County (FIPS 46113) until 2015, which resulted in some missing data. We manually renamed it in the data and extrapolated missing data from previous years.} which includes a county adjacency network, county-level demographic data (e.g., education, income, birth rates, USDA economic typology,\footnote{\url{https://www.ers.usda.gov/data-products/county-typology-codes/}}  and unemployment rates), and the Social Connectedness Index (SCI)~\citep{bailey2018social} measuring the relative frequency of Facebook friendships between each pair of counties. We aggregate all votes at the county level, treating each county as a chooser $a$ and using county features as $\bm{x}_a$ (modeling voting choices in aggregate is standard practice~\citep{alvarez1998politics}). For the graph $G$, we tested using both the geographic adjacency network and a network formed by connecting each county to the 10 others with which it has the highest SCI. We found almost identical results with both networks, so we only only discuss the results using the SCI network. We refer to the resulting dataset as \textsc{us-election-2016}.

\subsection{California precinct-level election data}

The presidential election data is particularly interesting because different ballots have different candidates, all running in the same election. For instance, this is analogous to having different regional availability of goods within a category in an online shopping service. In our next two datasets, \textsc{ca-election-2016} and \textsc{ca-election-2020}, we highlight a different scenario: when ballots in different locations may have different \emph{elections}. Extending the online shopping analogy, this emulates the case where different users view different recommended categories of items. Although it is beyond the scope of the present work, a discrete choice approach would enable measuring cross-election effects, such as coattail effects~\citep{hogan2005gubernatorial,ferejohn1984presidential} where higher-office elections increase excitement for down-ballot races.

To construct these datasets, we used data from the 2016 and 2020 California general elections from the Statewide Database.\footnote{\url{https://statewidedatabase.org}; 2016 and 2020 data accessed 8/20/20 and 3/22/21, resp.} This includes per-precinct registration and voting data as well as shapefiles describing the geographic boundaries of each precinct (California has over 20,000 voting precincts). The registration data contains precinct-level demographics (counts for party affiliation, sex, ethnicity, and age ranges), although such data was not available for all precincts. We restrict the data to the precincts for which all three data types were available: voting, registration, and shapefile (99.8\% of votes cast are included in our processed 2016 data, and 99.0\% in our 2020 data). Again, we treat each precinct as a chooser $a$ with demographic features $\bm{x}_a$.

Our processed California data includes elections for the US Senate, US House of Representatives, California State Senate, and California ballot propositions. We set aside presidential votes due to overlap with the previous dataset and state assembly votes to keep the data size manageable. Due to California's nonpartisan top-two primary system,\footnote{\url{https://www.sos.ca.gov/elections/primary-elections-california}} there are two candidates running for each office ---however, each voter has a different set of elections on their ballot due to differences in US congress and California state senate districts (the state has 53 congressional districts and 40 state senate districts). A discrete choice approach enables us to train a single model accounting for preferences over all types of candidates. We use the precinct adjacency network $G$ (since SCI is not available at the finer-grained precinct level), which we constructed from the Statewide Database shapefiles using QGIS (\url{https://qgis.org}).

\section{Empirical results}

We begin by demonstrating the sample complexity benefit of using network structure through Laplacian regularization on synthetic data. We then apply all three approaches to our datasets, compare their performance, and demonstrate the insights provided by a networked discrete choice approach. See \Cref{tab:datasets} in \Cref{sec:data} for a dataset overview. Our code and instructions for reproducing results are available at \url{https://github.com/tomlinsonk/graph-based-discrete-choice/}.

\subsection{Improved sample complexity with Laplacian regularization}\label{sec:sample-complexity}
\begin{figure}
\centering
  \includegraphics[width=0.7\columnwidth]{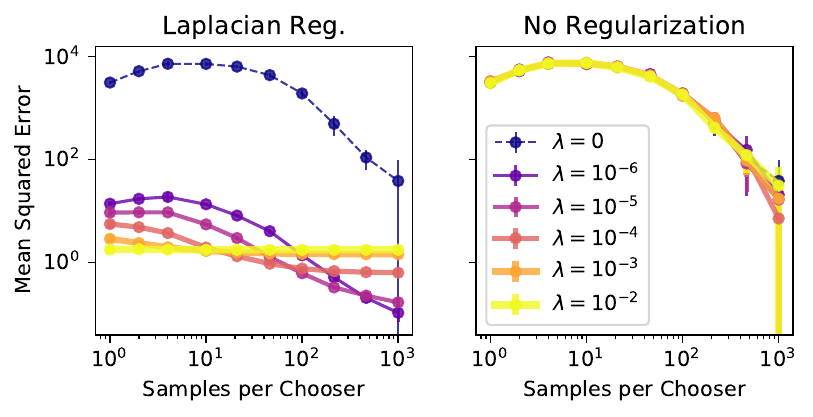}
  \caption{Estimation error of item utilities with (left) and without (right) Laplacian regularization on synthetic data generated according to the priors in \Cref{thm:bayesian}, with varying homophily strength $\lambda$. Error bars (most are tiny) show standard error over 8 trials. Using Laplacian regularization can improve sample complexity by orders of magnitude.}\label{fig:convergence_rate}
\end{figure}

By leveraging correlations between node preferences through Laplacian regularization, we need fewer samples per node in order to achieve the same inference quality. When preferences are smooth over the network, an observation of a choice by one node gives us information about the preferences of its neighbors (and its neighbors' neighbors, etc.), effectively increasing the usefulness of each observation.  In \Cref{fig:convergence_rate}, we show the sample complexity benefit of Laplacian regularization in synthetic data with 100-node Erd\H{o}s--R\'{e}nyi graphs ($p=0.1$) and preferences over 20 items generated according to the prior from \Cref{thm:bayesian}. In each of 8 trials, we generate the graph, sample utilities, and then simulate a varying number of choices by each chooser. We repeat this for different homophily strengths $\lambda$. For each simulated choice, we first draw a choice set size uniformly between 2 and 20, then pick a uniformly random choice set of that size. We then measure the mean-squared error in inferred utilities of observed items (fixing the utility of the first item to 0 for identification). When applying Laplacian regularization, we use the corresponding value of $\lambda$ used to generate the data (in real-world data, this needs to be selected through cross-validation). We train the models for 100 epochs. 

In this best-case scenario, we need orders of magnitude fewer samples per chooser if we take advantage of preference correlations: with Laplacian regularization, estimation error with only 1 sample per chooser is lower than the estimation error with no regularization and 1000 samples per chooser. The stronger the homophily, the fewer observations are needed to achieve optimal performance, since a node's neighbor's choices are more informative.

\subsection{Prediction performance comparison}

We now evaluate our approaches on real-world choice data. In the style of semi-supervised learning, we use a subset of choosers for training and held-out choosers for validation and testing. This emulates a scenario where it is too expensive to gather data from everyone in the network or existing data is not available for all nodes (e.g., perhaps not all individuals have consented to choice data collection). We vary the fraction of training choosers from 0.1 to 0.8 in increments of 0.1, using half of the remaining choosers for validation and half for testing. We perform 8 independent sampling trials at each fraction in the election datasets and 64 in the smaller Friends and Family datasets. 

As a baseline, we use standard logit models with no network information. For the election datasets, we use an MNL that uses county/precinct features to predict votes. This approach to modeling elections is common in political science~\citep{dow2004multinomial}. For \textsc{app-install}, we use a simple logit. For \textsc{app-usage}, we use a conditional logit (CL) with recency (as defined in \Cref{sec:app-data}). We then compare the three graph-based methods we propose to the baseline choice model: a GCN-augmented MNL (or CML), a Laplacian-regularized logit (or CL/MNL) with per-chooser utilities, and choice fraction propagation. Aside from propagation, we train the other methods with batch Rprop~\citep{riedmiller1993direct}, as implemented in PyTorch~\citep{paszke2019pytorch}. For each dataset--model pair, we select the hyperparameters that result in the lowest validation loss in a grid search; we tested learning rates $10^{-3}, 10^{-2}, 10^{-1}$ and $L_2$ regularization strengths $10^{-5}$, $10^{-4}$, $10^{-3}$, $10^{-2}$, $10^{-1}$ (we also tested no $L_2$ regularization in the two app datasets). We similarly select Laplacian $\lambda$ using validation data from $10^{-5}, 10^{-4}, 10^{-3}, 10^{-2}$ in the election datasets (in addition to these, we also test $10^0, 10^{-1}, 10^{-6}, 10^{-7}$ in the app datasets) and propagation $\rho$ from $0.1, 0.25, 0.5, 0.75, 1$. The smaller hyperparameter ranges in the election datasets were used due to runtime constraints. We train the likelihood-based models for 100 epochs, or until the squared gradient magnitude falls below $10^{-8}$. For propagation, we perform 256 iterations, breaking if the sum of squared differences between consecutive iterates falls below $10^{-8}$. We note that we did not aggressively fine-tune the GCN beyond learning rate and $L_2$ regularization strength, since it has many more hyperparameters than our other approaches and is more expensive to train. Our GCN results should therefore be interpreted as the performance a discrete choice practitioner should expect to achieve in a reasonable amount of time using the model, which we believe is an important metric.

\begin{figure*}
  \centering
  \setlength{\tabcolsep}{0.5mm}
  \begin{tabular}{lllll}
    \includegraphics[height=31mm]{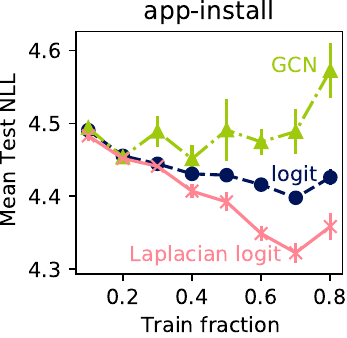} &
    \includegraphics[height=31mm]{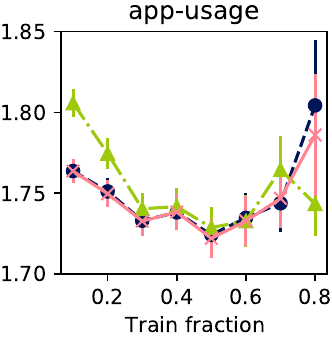} &
    \includegraphics[height=31mm]{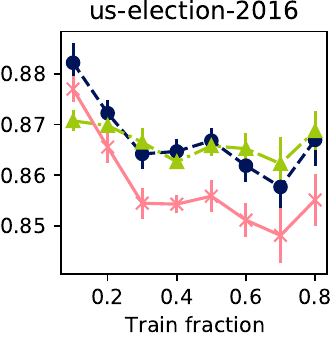} & 
\includegraphics[height=31mm]{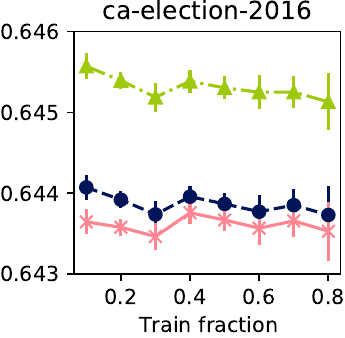} &
\includegraphics[height=31mm]{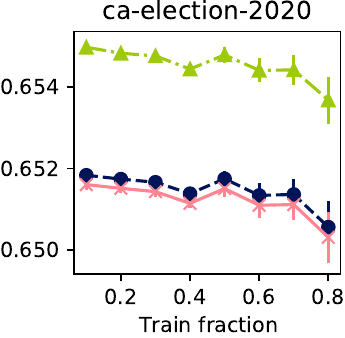}\\[-1.5mm]
  \includegraphics[height=31mm]{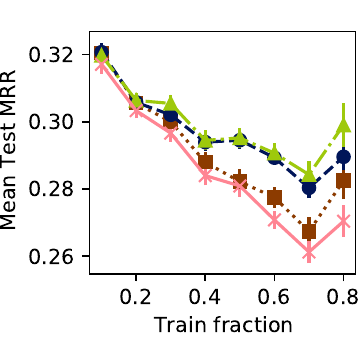} &
    \includegraphics[height=31mm]{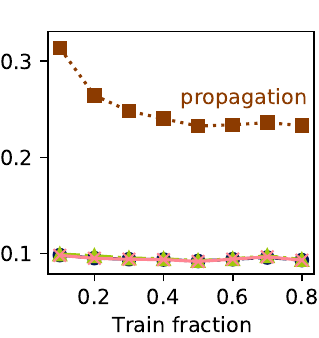} &
    \includegraphics[height=31mm]{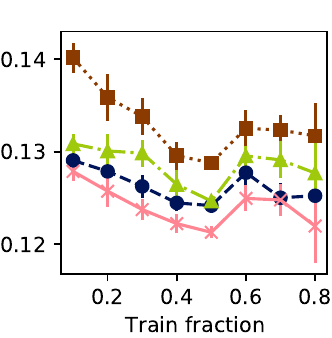} &
\includegraphics[height=31mm]{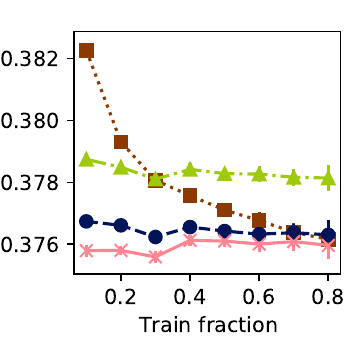} &
\includegraphics[height=31mm]{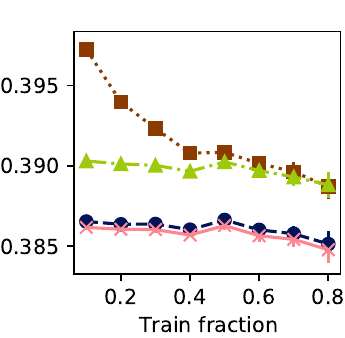}
  \end{tabular}
  \caption{Test negative log likelihoods (NLL; top row; lower is better) and mean relative ranks (MRR; bottom row; lower is better) on the two Friends and Family datasets and three election datasets (error bars show standard error over chooser sampling). ``Logit'' signifies plain logit in \textsc{app-install}, CL in \textsc{app-usage}, and MNL in the election datasets. Laplacian regularization improves performance in \textsc{app-install}, while no method improves on CL in \textsc{app-usage}. In the election data, Laplacian MNL, but not GCN, outperforms MNL across train fractions. Propagation performs well on \textsc{app-install}, but very poorly on \textsc{app-usage}, as it does not utilize recency.  Despite not using county/precinct features, propagation can be competitive in the election data.}\label{fig:prediction-results}
\end{figure*}

In \Cref{fig:prediction-results}, we show results of all four approaches on all five datasets. We evaluate the three likelihood-based methods using their test set negative log likelihood (NLL) and use \emph{mean relative rank} (MRR)~\citep{tomlinson2021learning} to evaluate propagation. For one sample, MRR is defined as the relative position of the actual choice in the list of predictions in decreasing confidence order (where 0 is the beginning of the list and and 1 is the end). We then report the mean MRR over the test set. In \textsc{app-install}, both Laplacian regularization and propagation improve prediction performance over the baseline logit model, and the advantage increases with the fraction of participants used for training (up to $6.8\%$ better MRR). However, the GCN performs worse than logit in terms of likelihood and the same or worse in terms of MRR. In contrast, graph-based methods do not outperform a conditional logit in \textsc{app-install}. In the three election datasets, Laplacian-regularized MNL consistently outperforms MNL (with up to $2.6\%$ better MRR in \textsc{us-election-2016}; the margin in the California data is small but outside errorbars), while the GCN performs on par with MNL in \textsc{us-election-2016} and worse in the California datasets. 

These results yield insight into the role networks play in different choice behaviors.  In \textsc{app-usage}, we find no benefit from using social network structure using any method. Instead, the recency feature appears to dominate, with propagation (which has no access to item features) performing much worse than the three  models that do incorporate recency. This indicates that app usage is driven by individual habit rather than by external social factors. On the other hand, our results show that app \emph{installation} has a strong social component: even simple Bluetooth proximity between friends provides a signal that they will install (but not necessarily use) similar apps. This finding highlights how combining a discrete choice approach with network data can illuminate the role social networks play in different choice behaviors. In the election data, especially \textsc{ca-election-2016}, even simple choice propagation performs remarkably well, despite \emph{entirely ignoring demographic features}. This reveals that many of the important predictive demographic features (such as party affiliation, age, and ethnicity) are so strongly correlated over the adjacency network that we don't need to know information about you to predict your vote: it suffices to know about your neighbors or your neighbors' neighbors.

\begin{table}
\centering
 \caption{Runtime in seconds to train and test each model, with standard err over 4 trials.}\label{tab:runtime}
  \begin{tabular}{lrrrr}
  \toprule
  Dataset & CL/MNL & Laplacian & GCN & Propagation\\ 
  \midrule
    \textsc{app-install} & $2.5 \pm 0.0$ & $2.2 \pm 0.0$ & $9.0 \pm 0.2$ & $0.2 \pm 0.0$ \\
  \textsc{app-usage} & $12 \pm 0$ & $13 \pm 0$ & $41 \pm 0$ & $0.9 \pm 0.0$ \\
      \textsc{us-election-2016} & $18 \pm 0$ & $19 \pm 0$ & $20 \pm 0.0$ & $1.0 \pm 0.0$ \\
    \textsc{ca-election-2016} & $605 \pm 6$ & $647 \pm 3$ & $758 \pm 4$ & $63 \pm 0$ \\
  \textsc{ca-election-2020} & $450 \pm 79$ & $397 \pm 2$ & $485 \pm 51$ & $38 \pm 0$ \\
  \bottomrule
  \end{tabular}

\end{table}

We also compare the runtime of each method. To measure runtime, each model was run on a 50-25-25 train-validation-test split of each dataset four times. Since the hyperparameters are not crucial for runtime measurements (especially because Rprop is not sensitive to initial learning rate as an adaptive method), we fixed the learning rate at 0.01, $L_2$ regularization strength at 0.001, Laplace regularization strength at 0.0001, and propagation $\rho$ at 0.5. For each trial, we trained and tested each model once, shuffling the order of models to avoid systematic bias due to caching. Laplacian regularization has very low overhead over CL/MNL, while GCN is up to $4\times$ slower in the smaller datasets (see \Cref{tab:runtime}). In the larger datasets, PyTorch's built-in parallelism reduces this relative gap. Propagation is more than $10\times$ faster than the choice models in every dataset.    

\subsection{Facebook and Myspace communities in \textsc{app-install}}
Given that we observed significant improvement in prediction performance in \textsc{app-install}, we take a closer look at the patterns learned by the Laplacian-regularized logit compared to the plain logit. In particular, the Facebook and Myspace apps were in the top 20 most-preferred apps under both models. Given that these were competitor apps at the time,\footnote{The dataset is from 2010; Facebook surpassed Myspace's popularity in the US in 2009.} we hypothesized that they might be popular among different groups of participants. This is exactly what we observe in the learned parameters of the Laplacian-regularized logit. Facebook and Myspace are in the top 10 highest-utility apps for 70 and 27 participants, respectively (out of 139 total; we refer to these sets as $F$ and $M$). Intriguingly, the overlap between $F$ and $M$ is only 3. Moreover, looking at the Bluetooth interaction network, we find the edge densities are more than twice as high within each of $F$ and $M$ than between them (\Cref{tab:edge-density}), indicating they are true communities in the social network. In short, the Laplacian-regularized logit learns about two separate subcommunities, one in which Facebook is popular and one in which Myspace is popular. 

\begin{table}
\centering
\centering
\caption{Edge densities within/between the groups preferring Facebook ($|F| = 70$) and Myspace ($|M| = 27$) in \textsc{app-install}. Left: including the 3 choosers in $F\cap M$. Right: excluding $F\cap M$. } \label{tab:edge-density}
  \begin{tabular}{lrr}
  \toprule
  & $F$ & $M$\\
  \midrule
  $F$ & $11.2\%$ & $4.9\%$\\
  $M$ & $4.9\%$ & $11.7\%$\\
    \hline
  \end{tabular}
  \quad \quad \quad
    \begin{tabular}{lrr}
  \toprule
  & $F$ & $M$\\
  \midrule
  $F$ & $11.3\%$ & $5.8\%$\\
  $M$ & $5.8\%$ & $12.0\%$\\
    \bottomrule
  \end{tabular}
\end{table}

\subsection{Counterfactuals in the 2016 US election}\label{sec:election-counterfactuals}

\begin{table}
\centering
\caption{Maximum likelihood 2016 election outcomes under our model under the three scenarios in \Cref{sec:election-counterfactuals}. We show mean vote shares (with 95\% confidence interval over trials) for the top three predicted candidates and differences in state outcomes between the counterfactual prediction and reality. C: Clinton, T: Trump, Outcome: Electoral College votes. T $\rightarrow$ C denotes that a state won by Trump goes for Clinton under the model. States abbreviated by postal code.}\label{tab:counterfactual-election}
  \begin{tabular}{lrrr}
  \toprule
  & Scenario 1 & Scenario 2 & Scenario 3\\
  \midrule
  C \% & $47.7\pm 0.1$ & $50.7\pm 0.1$  & $37.5 \pm 2.2$\\
  T \% & $46.3\pm 0.1$ & $49.3 \pm 0.1$ & $37.0 \pm 1.9$ \\
  T $\rightarrow$ C & --- & ---& PA \\
  C $\rightarrow$ T & 
    ME$^*$, MN, NV, NH 
  & 
    ME$^*$, MN, NV, NH 

   & MN, NV, NH \\
  Other & --- & --- & RI (``None'')\\
  Outcome & T 326, C 205&  T 326, C 205 & T 304, C 223\\
  \bottomrule
  \multicolumn{4}{l}{\footnotesize{$^*$Maine allocates Electoral College votes proportionally---we assume a 3-1 split.}}\\
  \end{tabular}
\end{table}

One of the powerful uses of discrete choice models is applying them to counterfactual scenarios to predict what might happen under different choice sets (e.g., in assortment optimization~\citep{rusmevichientong2010dynamic}). For instance, we can use our models to make predictions about election outcomes if different candidates had been on ballots in 2016. However, we begin this exploration with a warning: making predictions from observational data is subject to \emph{confounders}, unobserved factors that affected both who was on which ballot and how the states voted. For example, only Nevadans had the option to vote for ``None of these options,'' and Nevada is an outlier in a number of ways that are likely to impact voting, including its reliance on tourism, high level of diversity, and lack of income tax. This makes it less likely that the preferences of Nevadans for ``None of these options'' will neatly generalize to voters in other states. There are causal inference methods of managing confounding in discrete choice models; for instance, our county covariates act as regression controls~\citep{tomlinson2021choice}. If those covariates fully described variation in county voting preferences, then the resulting choice models would be unbiased, even with confounding~\citep{tomlinson2021choice}. However, we do not believe the covariates fully describe voting, since we can improve prediction by using regional or social correlations not captured by the county features. Nonetheless, examining our model's counterfactual predictions is still instructive, demonstrating an application of choice models, providing insight into the model's behavior, and motivating randomized experiments to test predictions about the effect of ballot changes. We note that the MNL we use obeys IIA, preventing relative preferences for candidates changing within a particular county when choice sets change. However, since states contain many counties, they are mixtures of MNLs (which can violate IIA), so their outcomes can change under the model. 

A widespread narrative of the 2016 election is that third-party candidates cost Clinton the election by disproportionately taking votes from her~\citep{guardian,rollcall}. To test this hypothesis, we examine three counterfactual scenarios: (Scenario 1) all ballots have five options: Clinton, Trump, Johnson, Stein, and McMullin; (Scenario 2) ballots only list Clinton and Trump, and (Scenario 3) ballots are as they were in 2016, but ``None of these candidates'' is added to every ballot. For each scenario, we take the best (validation-selected) Laplacian-regularized MNL trained on 80\% of counties from each of the 8 county sampling trials and average their vote count predictions. Maximum-likelihood outcomes under the model are shown in \Cref{tab:counterfactual-election}. We find no evidence to support the claim that third-party candidates hurt Clinton more than Trump. None of the scenarios changed the two major measures of outcome: Clinton maintained the popular vote advantage, while Trump carried the Electoral College. A few swing states change hands in the predictions. The model places more weight on ``None of these candidates'' than seems realistic (for instance, predicting it to be the plurality winner in Rhode Island), likely because training data is only available for this option in a single state, leading to confounding. We also note that under the true choice sets, the model's maximum likelihood state outcomes are the same as in Scenarios 1 and 2. A more complete analysis would examine the full distribution of Electoral College outcomes rather than just the maximum likelihood outcome, but we leave such analysis for future work as it is not our main focus.

\section{Discussion}
As we have seen, social and geographic network structure can be very useful in modeling the choices of a group of connected individuals, since people tend to have more similar preferences to their network neighborhood than to distant strangers. Several possible explanations are possible for this phenomenon: people may be more likely to become friends with similarly-minded individuals (homophily) or trends may spread across existing friendships (contagion). Unfortunately, determining  whether homophily or contagion is responsible for similar behavior among friends is notoriously difficult (and often impossible~\citep{shalizi2011homophily}).

We saw poor performance from the GCN relative to the logit models---as we noted, there are many hyperparameters that could be fine-tuned to possibly improve this performance, although this might not be practical for non-experts. Additionally, there are a host of other GNNs that could outperform GCNs in a choice task. Our contributions in this area are to demonstrate how GNN models can be adapted for networked choice problems and to encourage further exploration of such problems. However, our findings are consistent with several lines of recent work that show simple propagation-based methods outperforming graph neural networks~\citep{huang2020combining,wu2019simplifying,he2020lightgcn}. 

There are several interesting avenues for future work in graph-based methods for discrete choice. As we noted, much of the recent machine learning interest in discrete choice~ \citep{seshadri2019discovering,bower2020preference,rosenfeld2020predicting,tomlinson2021learning} has revolved around incorporating context effects (violations of IIA). Combining our methods with such approaches could answer questions thats are to our knowledge entirely unaddressed in the literature (and possibly even unasked): Do context effects have a social component? If so, what kinds of context effects? Can we improve contextual choice prediction with social structure (in terms of accuracy or sample complexity)? Another natural extension of our work is to use a weighted Laplacian when we have a weighted social network. Jointly modeling choices and network dynamics would also be very interesting, for instance using stochastic actor-oriented models~\cite{snijders1996stochastic,snijders2010introduction} for choice prediction. In another direction, choice data could be studied as an extra signal for community detection in networks, building on our identification of the Facebook and Myspace communities in the Friends and Family data.

\section*{Acknowledgments}
This research was supported by ARO MURI, ARO Award W911NF19-1-0057, NSF CAREER Award IIS-2045555, and NSF DMS-EPSRC Award 2146079. We thank Marios Papachristou for helpful discussions.

\bibliographystyle{abbrvnat}
\bibliography{references}

\end{document}